\documentclass[10pt,twocolumn,letterpaper]{article}

\usepackage{iccv}
\usepackage{times}
\usepackage{epsfig}
\usepackage{graphicx}
\usepackage{amsmath}
\usepackage{amssymb}
\usepackage{mathtools}
\usepackage{graphics}
\usepackage{adjustbox}
\usepackage{stmaryrd}
\usepackage{tabu}
\usepackage{nth}
\usepackage{multirow}
\usepackage{bm}
\usepackage{textcomp}

\usepackage[breaklinks=true,bookmarks=false]{hyperref}

\iccvfinalcopy 

\def\iccvPaperID{27} 

\ificcvfinal\fi

\begin{document}

\title{An audiovisual and contextual approach for categorical and continuous emotion recognition in-the-wild}

\author{Panagiotis Antoniadis\thanks{Equal contribution} , Ioannis Pikoulis$^*$, Panagiotis P. Filntisis, Petros Maragos \\
\normalsize{School of ECE, National Technical University of Athens, 15773 Athens, Greece}\\
{\texttt{\small
\{pantoniadis97, pikoulis.giannis\}@gmail.com, filby@central.ntua.gr, maragos@cs.ntua.gr
}}
}


\maketitle
\ificcvfinal\thispagestyle{empty}\fi

\begin{abstract}
In this work we tackle the task of video-based audio-visual emotion recognition, within the premises of the 2nd Workshop and Competition on Affective Behavior Analysis in-the-wild (ABAW2). Poor illumination conditions, head/body orientation and low image resolution constitute factors that can potentially hinder performance in case of methodologies that solely rely on the extraction and analysis of facial features. In order to alleviate this problem, we leverage both bodily and contextual features, as part of a broader emotion recognition framework. We choose to use a standard CNN-RNN cascade as the backbone of our proposed model for sequence-to-sequence (seq2seq) learning. Apart from learning through the RGB input modality, we construct an aural stream which operates on sequences of extracted mel-spectrograms. Our extensive experiments on the challenging and newly assembled Aff-Wild2 dataset verify the validity of our intuitive multi-stream and multi-modal approach towards emotion recognition ``in-the-wild''. Emphasis is being laid on the the beneficial influence of the human body and scene context, as aspects of the emotion recognition process that have been left relatively unexplored up to this point. All the code was implemented using \mbox{PyTorch}\footnote{\url{https://pytorch.org/}} and is publicly available\footnote{\url{https://github.com/PanosAntoniadis/NTUA-ABAW2021}}.
\end{abstract}

\section{Introduction}
Automatic affect recognition constitutes a subject of rigorous studies across several scientific disciplines and bears immense practical importance as it has extensive applications in environments that involve human-robot cooperation, sociable robotics, medical treatment, psychiatric patient surveillance and many other human-computer interaction scenarios.

Representing human emotions has been a basic topic of research in psychology. While the cultural and ethnic background of a person can affect their expressive style, Ekman indicated that humans perceive certain basic emotions in the same way regardless of their culture \cite{ekman1971constants, ekman1994strong}. These six universal facial expressions (happiness, sadness, surprise, fear, disgust and anger) constitute the categorical model. Contempt was subsequently added as one of the basic emotions \cite{matsumoto1992more}. Due to its direct and intuitive definition of facial expressions, the categorical model is used in the majority of emotion recognition algorithms \cite{jung2015joint, hasani2017facial, zhao2016peak, antoniadis2021exploiting} and large-scale databases (MMI \cite{pantic2005web}, AFEW \cite{dhall2013emotion}, FER-Wild \cite{mollahosseini2016facial}, etc). However, the subjectivity and ambiguity of restricting human emotion to discrete categories result in large intra-class variations and small inter-class differences. 

Recently, the dimensional model proposed by Russell \cite{russell1980circumplex} has gained a lot of attention where emotion is described using a set of two latent dimensions that are valence (how pleasant or unpleasant a feeling is) and arousal (how likely is the person to take action under the emotional state). Another dimension called dominance is used sometimes to know whether the person is controlling the situation or not. Since a continuous representation can distinguish between subtly different displays of affect and encode small changes in the intensity, some recent algorithms \cite{nicolaou2011continuous, chang2017fatauva, kollias2019expression} and databases (Aff-Wild \cite{zafeiriou2017aff, kollias2019deep}, Aff-Wild2 \cite{kollias2019expression}, OMG-Emotion \cite{barros2018omg}, AFEW-VA \cite{kossaifi2017afew}, etc) have utilized the dimensional model for uncontrolled emotion recognition. Even so, predicting a 2-dimensional continuous value instead of a category increases the task complexity by a lot and lacks intuitiveness. 

The remainder of the paper is structured as follows: Firstly, we provide an overview of the latest and most notable related work in the domain of video-based emotion recognition in-the-wild. Subsequently, we analyze our proposed model architecture. Next, we present our experimental results on the Aff-Wild2 dataset, followed by conclusive remarks.

\section{Related Work}

Emotion recognition has been extensively studied for many years using different representations of human emotion, like basic facial expressions, action units and valence-arousal. 

Recently, many studies have tried to leverage all emotion representations and jointly learn these three facial behavior tasks. Kollias et al. \cite{kollias2019face} proposed the FaceBehaviorNet, the first study that considered joint-learning of all facial behaviour tasks, in a single holistic framework. They utilized many publicly available emotion databases and proposed two strategies for coupling the tasks during training. Later, Kollias et al. released the Aff-Wild2 dataset \cite{kollias2019expression}, the first large scale in-the-wild database containing annotations for all three main behavior tasks. They also proposed  multitask learning models that employ both visual and audio modalities and suggested using the ArcFace loss \cite{deng2019arcface} for expression recognition. In an additional work, Kollias et al. \cite{kollias2021affect, kollias2021distribution} studied the problem of non-overlapping annotations in multitask learning datasets. They explored task-relatedness and proposed a novel distribution matching approach, in which knowledge exchange is enabled between tasks, via matching of their predictions’ distributions. Last year, the First Affective Behavior Analysis in-the-wild (ABAW) Competition \cite{kollias2020analysing} was held in conjunction with the IEEE Conference on Face and Gesture Recognition 2020. The competition contributed in advancing the state-of-the-art methods on the dimensional, categorical and facial action unit analysis and recognition on the basis of the Aff-Wild2 dataset.

Although, the aforementioned methodologies boast relatively high recognition scores on the premise of facial expression analysis, they often neglect the usage of other supplementary sources of affective information, such as the \textit{body} and \textit{context}. Related works \cite{kosti2019context, mittal2020emoticon, NTUA_BEEU, pikoulis2021leveraging} from the field of context-based visual emotion recognition, follow a more holistic approach towards solving the ``in-the-wild" version of the current problem. Kosti et al. \cite{kosti2019context} introduced the EMotions In Context (EMOTIC) dataset, the first large-scale image database for context-based emotion recognition, annotated on the basis of an extended emotional corpus with 26 discrete categories and VAD \mbox{dimensions}. \mbox{A baseline} model was also provided, consisting of two ConvNet feature extractors (one for each of the \textit{body} and \textit{context} input streams) and one fusion network. Mittal et al. \cite{mittal2020emoticon} surpassed baseline performance by fusing multiple input modalities, including the face, pose, inter-agent interactions and socio-dynamic context, effectively forming the EmotiCon framework. Furthermore, researchers have extended the concept of context-based visual emotion recognition in the dynamic setting of video sequences. Filntisis et al. \cite{NTUA_BEEU} improved upon the baseline performance of \cite{Luo2019ARBEETA} relative to the Body Language Dataset (BoLD) by incorporating a contextual feature encoding branch and a visual-semantic embedding loss based on Global Vectors (GloVe) \cite{pennington2014glove} word embeddings. Lastly, Pikoulis et al. \cite{pikoulis2021leveraging} achieved state-of-the-art performance on BoLD through an intuitive approach that included the use of scene and attribute characteristics as well as multi-stream optical flow.

\section{Method}

\begin{figure*}[t]
    \centering
    \includegraphics[width=.99\textwidth]{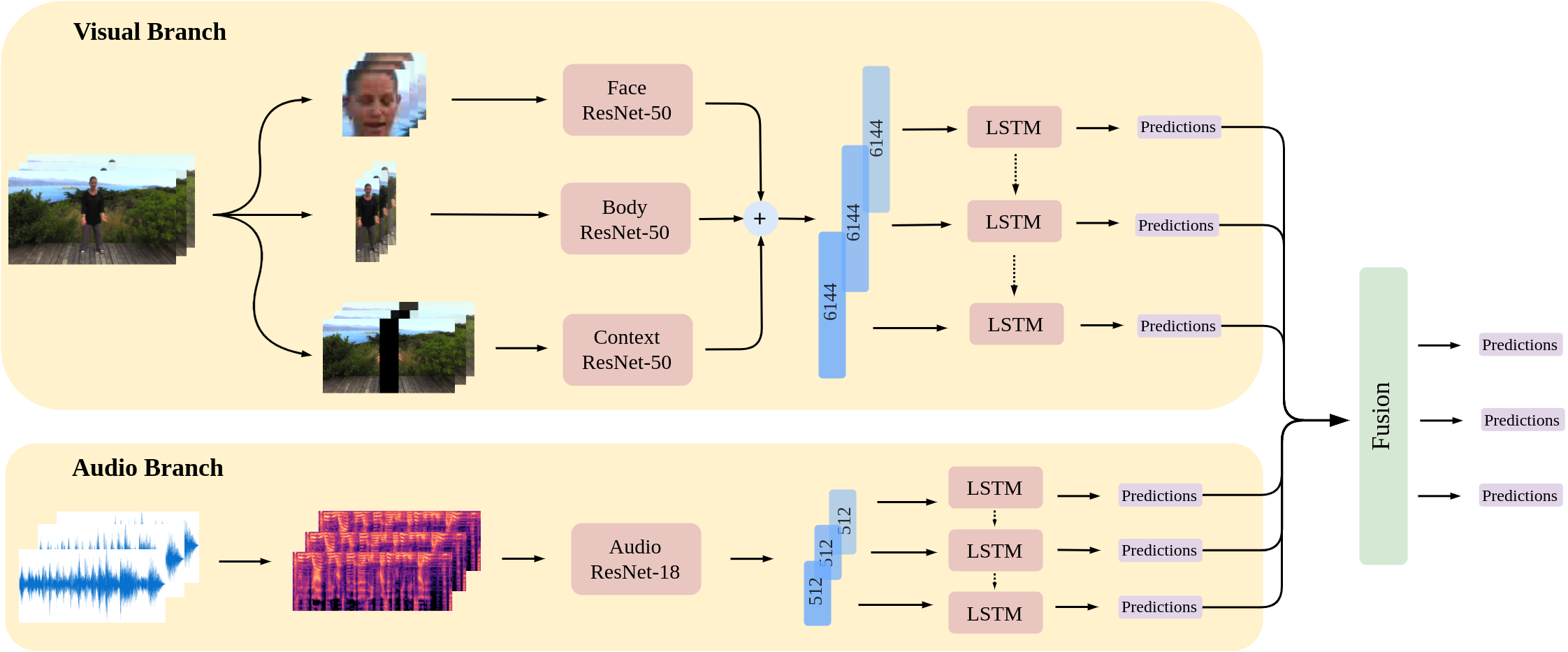}
    \caption{Overview of our proposed method.}
    \label{fig:overview}
\end{figure*}

A complete schematic diagram of our proposed model is shown in Fig. \ref{fig:overview}. Firstly, we will present the structure of the sub-network regarding the RGB visual modality, along with our proposed extensions for the enhancement of emotion understanding. Next, we will do the same for the aural model, and finally we will present the unified audio-visual architecture. 

\subsection{Seq2Seq}
In order to leverage temporal information and emotion labels throughout each video, our method takes as input sequences of frames that contain either visual or aural information (which will be described in the next subsections). After extracting intermediate feature representations for each sequence of frames, we use a standard LSTM (Long Short-Term Memory) \cite{hochreiter1997long} model in order to map the extracted feature sequences features to their respective output labels. 

\subsection{Visual Model}
A single RGB image usually encodes static appearance at a specific point in time but lacks the contextual information about previous and next frames. We aspire to enhance the descriptive capacity of the extracted deep visual embeddings through the feature level combination of multiple feature extractors that focus on different parts of the human instance. In our implementation, we utilize the human face as our primary source of affective information, while we also make use of the body and surrounding depicted environment, in a supplementary manner. For all of the subsequent convolutional branches, we use a standard 50-layer ResNet \cite{he2016deep}, as our feature extractor backbone. The ResNet-50 variant produces 2048-dim deep feature vector representations for each given input image. All \mbox{ConvNet} backbones are pre-trained using various task-specific datasets. As it will be discussed later on, pre-training constitutes the main differentiating factor among the multiple ConvNet backbones that comprise our visual model. 

\subsubsection{Face}
The face is commonly perceived as the window to the human soul and the most expressive source of visual affective information. We introduce an input stream which explicitly operates on the aligned face crops of the primary depicted human agents. The localization, extraction and alignment of face regions per frame has been carried out by the official distributors of the Aff-Wild2 dataset priorly. During frames where face detection and alignment has failed and the corresponding face crops are missing, we feed the \mbox{ConvNet} feature extractor with an input tensor of appropriate size, filled with zeros.

The \mbox{ConvNet} backbone of the face branch receives manual pre-training on AffectNet \cite{mollahosseini2017affectnet} which constitutes the largest facial expression database, containing over 1M images, annotated on both categorical and dimensional level. We pre-trained the face branch on AffectNet for 5 epochs using a batch size of 64 and a learning rate of 0.001 achieving 64.09\% validation accuracy. 

\subsubsection{Context}
We incorporate a context stream in the form of RGB frames whose primary depicted agents have been masked out. For the acquisition of the masks we use body bounding boxes and multiply them element-wise with a constant value of zero. Prior to the acquisition of the body bounding boxes and the corresponding masks, we calculate the 2D coordinates for 25 joints of the body of the primary depicted agent using the BODY25 model of the OpenPose \cite{cao2018openpose} publicly available toolkit. Let $\{({x^{(t)}_n}, {y^{(t)}_n})\}_{n=1}^{25}$ be the detected set of horizontal and vertical joint coordinates of a given agent, at frame $t$. The bounding box of the agent $\textrm{bbox}_{\textrm{agent}}$ within a given image $I$ of height $H$ and width $W$, is calculated as follows: 
\begin{equation}
\resizebox{.9\hsize}{!}{
$\begin{gathered}
    e_{z} = \lambda_{z}(\max_{n}{z^{(t)}_{n}}-\min_{n}{z^{(t)}_{n}}), \quad z\equiv x \mbox{ } \textrm{or} \mbox{ } z\equiv y\\ 
    \textrm{bottom}=\min{(\max_{n}{y^{(t)}_{n}}+e_{y}, H)}, \quad \textrm{right}=\min{(\max_{n}{x^{(t)}_{n}}+e_{x}, W)} \\
    \textrm{top}=\max{(0, \min_{n}{y^{(t)}_{n}}-e_{y})}, \quad \textrm{left}=\max{(0, \min_{n}{x^{(t)}_{n}}-e_{x})} \\
    \textrm{bbox}_{\textrm{agent}}=I(\textrm{top:}\textrm{bottom}, \textrm{left:}\textrm{right})
\end{gathered}$}
\end{equation}
All joints with a detection confidence score that is less than 10\%, are discarded. The masked image $I_{\textrm{mask}}$ for a given input image $I$ is calculated as follows: 
\begin{equation}
    I_{\textrm{mask}}=\begin{cases}
    I(i,j) &\mbox{if } (i,j)\notin{\textrm{bbox}_{\textrm{agent}}} \\
    0 &\mbox{otherwise }
    \end{cases}
\end{equation}
where the tuple $(i,j)$ corresponds to all valid pixel locations. Throughout our experiments we set $\lambda_{x}=0.1$ and \mbox{$\lambda_{y}=0.25$}. 

Contextual feature extraction is a scene-centric task, and therefore we choose to initialize the corresponding \mbox{ConvNet} backbone using the Places365-Standard \cite{zhou2016places}, a large-scale database of photographs, labeled with scene \mbox{semantic} \mbox{categories}. The pre-trained model is publicly available\footnote{\url{https://github.com/CSAILVision/places365}}.

\subsubsection{Body}
We incorporate an additional input stream that focuses solely on encoding bodily expressions of emotion, with the aim of alleviating the problem of undetected or misaligned face crops. The contribution of the body branch in the emotion recognition process becomes more evident during frames where the corresponding face crops are not existent. The body stream operates either on the bounding of the depicted agents or the entire image. The ConvNet feature extractor is pre-trained on the object-centric ImageNet \cite{deng2009imagenet} database. The pre-trained weights are publicly available\footnote{\url{https://download.pytorch.org/models/resnet50-0676ba61.pth}}.   

The early feature fusion of all of the aforementioned input streams results in a 6144-dim concatenated feature vector. Subsequently, the fused features are fed into a bidirectional, single-layer LSTM, with 512 hidden units, for further temporal modeling.

\subsection{Aural Model}
In the branch that incorporates audio information, starting with a sequence of input waveforms, we extract the mel-spectrogram representation of each. Then, we use a 18-layer ResNet model pretrained on the ImageNet dataset to extract a 512 feature vector for each input waveform. Finally, using an LSTM layer, we map the feature sequence to labels (either expression labels in the case of Track 2 or VA labels in the case of Track 1). 

\subsection{Loss Functions}
For Track 1 of the ABAW competition, we use both a standard \textit{mean-squared error} $\mathcal{L}_{\textrm{mse}}$ as well as loss term based on the \textit{concordance correlation coefficient} (CCC). The latter is defined as:

\begin{equation}
    \rho_c = \frac{2 s_{xy}}{s_x^2 + s_y^2 + (\bar{x} - \bar{y})^2}
\end{equation}
where $s_x$ and $s_y$ denote the variance of the predicted and ground truth values respectively, $\bar{x}$ and $\bar{y}$ are the corresponding mean values and $s_{xy}$ is the respective covariance value. The range of CCC is from -1 (perfect disagreement) to 1 (perfect agreement). Hence, in our case we define $\mathcal{L}_{\textrm{ccc}}$ as:

\begin{equation}
    \mathcal{L}_{\textrm{ccc}} = 1 - \frac{\rho_v + \rho_a}{2}
    \label{eqn:ccc}
\end{equation}
where $\rho_v$ and $\rho_a$ are the respective CCC of valence and arousal.

For Track 2, we use a standard cross-entropy function $\mathcal{L}_{\textrm{ce}}$. We also enforce semantic congruity between the extracted visual embeddings and the categorical label word embeddings from a 300-dim GloVe \cite{pennington2014glove} model, pre-trained on Wikipedia and Gigaword 5 data, in the same manner as in \cite{NTUA_BEEU}. More specifically, given an input sample $\bm{\mathbf{x}}$, we transform the concatenated visual embeddings $f_{v}(\bm{\mathbf{x}})$ into the same dimensionality as the word embeddings $f_{t}(y)$ through a linear transformation $\bm{\mathbf{W}}_{\textrm{emb}}$, with $y$ being the ground truth target label. We later apply an MSE loss between the transformed visual embeddings and the word embeddings which correspond to the ground truth emotional label and denote this term as $\mathcal{L}_{\textrm{emb}}$:
\begin{equation}
    \mathcal{L}_{\textrm{emb}}=\big\|\bm{\mathbf{W}}_{\textrm{emb}}f_{v}(\bm{\mathbf{x}})-\sum_{c}\llbracket y=c\rrbracket f_{t}(y)\big\|_{2}^{2}
\end{equation}
\noindent where $\llbracket \cdot \rrbracket$ is the Iverson bracket and $c$ is the class index. The whole network can be trained in an end-to-end manner by minimizing the combined loss function \mbox{$\mathcal{L}=\mathcal{L}_{\textrm{ce}}+\lambda_{\textrm{emb}}\mathcal{L}_{\textrm{emb}}$}. For simplicity, we set $\lambda_{\textrm{emb}}=1.0$ during all of our subsequent experiments.           

\section{Experimental Results}
\subsection{Validation Set}
Tables \ref{tab:expr} and \ref{tab:va} present our results on the Aff-Wild2 validation set, for the Expression and Valence-Arousal sub-challenges respectively, together with a performance comparison with the baseline and top entries from last year's ABAW competition \cite{kollias2020analysing}.
\begin{table}[ht!]
\centering
\caption{Results on the Aff-Wild2 validation set, for the Expression sub-challenge.}
\vspace{0.1cm}
\resizebox{8.35cm}{!}{
\begin{tabular}{cccc}
\hline
Method                   & $F_1$ Score       & Accuracy       & Total          \\ \hline\hline
Baseline \cite{kollias2021analysing}     & 0.30           & 0.50           & 0.366          \\ \hline
NISL2020 \cite{deng2020multitask}                          & -          & -         & 0.493 \\
ICT-VIPL \cite{liu2020emotion}                          & 0.33           & 0.64          & 0.434 \\
TNT \cite{kuhnke2020two}                      & -              & -              & 0.546          \\ \hline
Audio                    & 0.375          & 0.495          & 0.415          \\ 
Visual (F)               & 0.453          & 0.584          & 0.496          \\ 
Visual (BCF)             & 0.517          & \textbf{0.640} & 0.558          \\ 
Visual (BCF) + $\mathcal{L}_{\textrm{emb}}$      & \textbf{0.532} & 0.639          & \textbf{0.567} \\ \hline
Visual (F + BCF)         & 0.543          & 0.657          & 0.580          \\ 
Audio + Visual (F)       & 0.519          & 0.645          & 0.561          \\ 
Audio + Visual (BCF)     & 0.536          & 0.654          & 0.575          \\ 
Audio + Visual (F + BCF) & \textbf{0.555} & \textbf{0.668} & \textbf{0.592} \\ \hline
\end{tabular}}
\label{tab:expr}
\end{table}

\begin{table}[ht!]
\centering
\caption{Results on the Aff-Wild2 validation set, for the Valence-Arousal sub-challenge.}
\vspace{0.1cm}
\resizebox{8.35cm}{!}{
\begin{tabular}{cccc}
\hline
Method                   & CCC-V          & CCC-A         & Total          \\ \hline\hline
Baseline \cite{kollias2021analysing}                & 0.23           & 0.21           & 0.22           \\ \hline
NISL2020 \cite{deng2020multitask}                          & 0.335          & 0.515         & 0.425 \\
ICT-VIPL \cite{zhang2020m}                 & 0.32           & 0.55           & 0.435           \\
TNT \cite{kuhnke2020two}                      & 0.493          & 0.613          & 0.553          \\ \hline
Audio                    & 0.243          & 0.400          & 0.322          \\ 
Visual (F)               & 0.330          & 0.539          & 0.435          \\ 
Visual (BCF)             & \textbf{0.344} & \textbf{0.550} & \textbf{0.447} \\ \hline
Visual (F + BCF)         & 0.358          & 0.597          & 0.478          \\ 
Audio + Visual (F)       & 0.366          & 0.582          & 0.474          \\ 
Audio + Visual (BCF)     & 0.382          & 0.586          & 0.484          \\ 
Audio + Visual (F + BCF) & \textbf{0.386} & \textbf{0.616} & \textbf{0.502} \\ \hline
\end{tabular}}
\label{tab:va}
\end{table}

\begin{table*}[!t]
\centering
\caption{Results on the Aff-Wild2 test set relative to baseline methodologies, the top entries of the ABAW2 competition and our proposed models.}
\vspace{0.1cm}
\resizebox{17.4cm}{!}{
\begin{tabular}{cccccccccc}
\hline 
\multirow{2}{*}{Method}    & \multicolumn{3}{c}{Track 1}                      & \multicolumn{3}{c}{Track 2}                      & \multicolumn{3}{c}{Track 3}                      \\ \cline{2-10} 
                           & CCC-V          & CCC-A          & Total          & F1 Score       & Accuracy       & Total          & F1 Score       & Accuracy       & Total          \\ \hline \hline
MT-VGG  \cite{kollias2021affect}                   & 0.43           & 0.42           & 0.425          & -              & -              & 0.50           & -              & -              & 0.60           \\
V-MT-VGG-GRU \cite{kollias2021affect}               & 0.44           & 0.51           & 0.475          & -              & -              & 0.51           & -              & -              & 0.62           \\
A-MT-VGG-GRU \cite{kollias2021affect}               & 0.46           & 0.45           & 0.455          & -              & -              & 0.52           & -              & -              & 0.62           \\
A/V-MT-VGG-GRU \cite{kollias2021affect}             & 0.47           & 0.52           & 0.495          & -              & -              & 0.53           & -              & -              & 0.63           \\ \hline
NISL-2021 \cite{deng2020multitask}                & \textbf{0.533} & \textbf{0.454} & \textbf{0.494} & 0.431          & 0.654          & 0.505          & 0.451          & 0.847          & 0.653          \\
Maybe Next Time \cite{thinh2021emotion}              & -              & -              & -              & 0.605          & 0.729          & 0.646          & 0.461          & 0.877          & 0.669          \\
CPIC-DIR2021 \cite{jin2021multi}               & -              & -              & -              & 0.683          & 0.771          & 0.712          & 0.489          & \textbf{0.892} & 0.690          \\
Netease Fuxi Virtual Human \cite{zhang2021prior} & 0.486          & 0.495          & 0.490          & \textbf{0.763} & \textbf{0.807} & \textbf{0.778} & \textbf{0.506} & 0.888          & \textbf{0.697} \\ \hline
Visual (F)                 & \textbf{0.379} & 0.341          & 0.36           & 0.312          & 0.593          & 0.405          & -              & -              & -              \\
Visual (BCF)               & 0.316          & 0.352          & 0.334          & 0.329          & 0.611          & 0.423          & -              & -              & -              \\
Audio + Visual (F + BCF)   & 0.368          & \textbf{0.464} & \textbf{0.416} & \textbf{0.337} & \textbf{0.642} & \textbf{0.437} & -              & -              & -              \\ \hline
\end{tabular}}
\label{tab:test}
\end{table*}

The letter \textit{F} denotes a visual branch trained using only the cropped face in the input video, while \textit{BCF} denotes the visual branch that incorporates both bodily as well as contextual features, as seen in Figure~\ref{fig:overview}. \textit{Audio} denotes the results of the aural branch. In both tables we also include the results of the weighted average late score fusion among all different combinations of \textit{F}, \textit{BCF}, and \textit{Audio}. We readily see that including the body and the context as additional, supplementary information results in a significant performance boost, especially in the case of the Expression sub-challenge. Furthermore, the fusion of any two methods results in increased performance, when compared to the single branches. \mbox{Finally}, in both Tables, the fusion of all three different models results in the best scores, i.e. \textbf{0.592} for the Expression sub-challenge and \textbf{0.502} for the VA sub-challenge, on the basis of the Aff-Wild2 validation set. 

\subsection{Test Set}

Table \ref{tab:test} presents an extensive performance comparison among the proposed methodologies of \cite{kollias2021affect}, the top entries of this year's ABAW2 competition as well as our own \mbox{models}. 

It is quite evident that our proposed model, even though it manages to surpass the baseline \cite{kollias2021analysing} on the validation set by a large margin, it does not come close to the performance, exhibited by the top submissions \cite{deng2020multitask, thinh2021emotion, jin2021multi, zhang2021prior}. There two main reasons that account for this observed behavior. Firstly, we presume that the distribution of the test set greatly differs from that of the validation set, leading to a significant difference in performance, relative to the two sets. Furthermore, we followed a more holistic and high-level approach to the problem, whereas the aforementioned teams laid their emphasis explicitly on the facial stream and disregarded all other sources of affective information, such as the \textit{body}, \textit{context} or even the \textit{aural} modality. In that way, the production of the best facial expression model basically lies outside the scope of the current analysis. However, through our experiments we verified the beneficial influence of both the \textit{body} and \textit{context}, as supplementary streams that encode valuable information, especially in the case when the applied face extraction and alignment algorithms fail to due to extreme head or body orientations, low illumination and image quality. 

More specifically, on the Expression sub-challenge, the sole usage of the facial stream, led to a total score of 0.405, while the inclusion of both bodily and contextual features raised the total score to 0.423. With the introduction of the aural stream, our model reached a maximum of 0.437 on the test set. On the VA sub-challenge we notice a slightly different but similar behavior. The introduction of contextual and bodily features decreased the performance of our model significantly along the valence dimension but increased it along the arousal dimension. This can be justified by the fact that the latter is highly correlated with body motion while the former is dominated by facial expressions and AUs. The inclusion of the aural stream, once again led to a significant boost in overall performance, resulting in more than 10\% increase in CCC-A and a total score of 0.416.

\section{Conclusion}
We presented our method for the ``2nd Workshop  and  Competition  on  Affective  Behavior  Analysis in-the-wild  (ABAW2)" challenge. Apart from using only face-cropped images, we leverage both contextual (scene), bodily features, as well as the audio of the video, to further enhance our model's perception. Our results show that fusion of the different streams of information results in significant performance increase, when compared to both single streams, as well as previous best published results on the validation set. Even though our model does not come close to the performance of this year's top entries, we have successfully highlighted the beneficial influence of the \textit{body} and \textit{context} input streams in the emotion recognition process, and as a result, propose our multi-stream and multi-modal approach as a potential way to further improve methodologies that solely rely on the analysis of facial expressions. 

{\small
\bibliographystyle{ieee_fullname}
\bibliography{egbib}

\begin{thebibliography}{10}\itemsep=-1pt

\bibitem{antoniadis2021exploiting}
Panagiotis Antoniadis, Panagiotis~P Filntisis, and Petros Maragos.
\newblock Exploiting emotional dependencies with graph convolutional networks
  for facial expression recognition.
\newblock {\em arXiv preprint arXiv:2106.03487}, 2021.

\bibitem{barros2018omg}
Pablo Barros, Nikhil Churamani, Egor Lakomkin, Henrique Siqueira, Alexander
  Sutherland, and Stefan Wermter.
\newblock The {OMG-Emotion} behavior dataset.
\newblock In {\em IEEE Int. Joint Conf. on Neural Networks (IJCNN)}, 2018.

\bibitem{cao2018openpose}
Z. {Cao}, G. {Hidalgo}, T. {Simon}, S.-E. {Wei}, and Y. {Sheikh}.
\newblock Open{P}ose: {R}ealtime multi-person 2{D} pose estimation using part
  affinity fields.
\newblock {\em IEEE Trans. on Pattern Analysis and Machine Intelligence
  (TPAMI)}, 43:172--186, 2021.

\bibitem{chang2017fatauva}
Wei-Yi Chang, Shih-Huan Hsu, and Jen-Hsien Chien.
\newblock {FATAUVA-Net}: An integrated deep learning framework for facial
  attribute recognition, action unit detection, and valence-arousal estimation.
\newblock In {\em Proc. IEEE Conf. on Computer Vision and Pattern Recognition
  Workshops (CVPRW)}, 2017.

\bibitem{deng2020multitask}
Didan Deng, Zhaokang Chen, and Bertram~E Shi.
\newblock Multitask emotion recognition with incomplete labels.
\newblock In {\em 15th IEEE Int. Conf. on Automatic Face and Gesture
  Recognition (FG)}. IEEE, 2020.

\bibitem{deng2009imagenet}
J. Deng, W. Dong, R. Socher, L.-J. Li, K. Li, and L. Fei-Fei.
\newblock Image{N}et: A large-scale hierarchical image database.
\newblock In {\em {IEEE Conf. Computer Vision and Pattern Recognition (CVPR)}}.
  IEEE, 2009.

\bibitem{deng2019arcface}
Jiankang Deng, Jia Guo, Niannan Xue, and Stefanos Zafeiriou.
\newblock {ArcFace}: Additive angular margin loss for deep face recognition.
\newblock In {\em Proc. of the IEEE/CVF Conf. on Computer Vision and Pattern
  Recognition (CVPR)}, 2019.

\bibitem{dhall2013emotion}
Abhinav Dhall, Roland Goecke, Jyoti Joshi, Michael Wagner, and Tom Gedeon.
\newblock Emotion recognition in the wild challenge 2013.
\newblock In {\em Proc. 15th ACM Int. Conf. on Multimodal Interaction}, 2013.

\bibitem{ekman1994strong}
Paul Ekman.
\newblock Strong evidence for universals in facial expressions: A reply to
  {Russell's} mistaken critique.
\newblock {\em Psychological bulletin}, 1994.

\bibitem{ekman1971constants}
Paul Ekman and Wallace~V Friesen.
\newblock Constants across cultures in the face and emotion.
\newblock {\em Journal of Personality and Social Psychology}, 17:124, 1971.

\bibitem{NTUA_BEEU}
P.~P. Filntisis, N. Efthymiou, G. Potamianos, and P. Maragos.
\newblock Emotion understanding in videos through body, context, and
  visual-semantic embedding loss.
\newblock In {\em {Proc. 16th Eur. Conf. Computer Vision Workshops (ECCVW) -
  Workshop on Bodily Expressed Emotion Understanding (BEEU)}}, 2020.

\bibitem{hasani2017facial}
Behzad Hasani and Mohammad~H Mahoor.
\newblock Facial expression recognition using enhanced deep {3D} convolutional
  neural networks.
\newblock In {\em Proc. IEEE Conf on Computer Vision and Pattern Recognition
  Workshops (CVPRW)}, 2017.

\bibitem{he2016deep}
K. He, X. Zhang, S. Ren, and J. Sun.
\newblock Deep residual learning for image recognition.
\newblock In {\em {Proc. IEEE Conf. Computer Vision and Pattern Recognition
  (CVPR)}}, 2016.

\bibitem{hochreiter1997long}
Sepp Hochreiter and J{\"u}rgen Schmidhuber.
\newblock Long short-term memory.
\newblock {\em Neural Computation}, 9(8):1735--1780, 1997.

\bibitem{jin2021multi}
Yue Jin, Tianqing Zheng, Chao Gao, and Guoqiang Xu.
\newblock A multi-modal and multi-task learning method for action unit and
  expression recognition.
\newblock {\em arXiv preprint arXiv:2107.04187}, 2021.

\bibitem{jung2015joint}
Heechul Jung, Sihaeng Lee, Junho Yim, Sunjeong Park, and Junmo Kim.
\newblock Joint fine-tuning in deep neural networks for facial expression
  recognition.
\newblock In {\em Proc. IEEE Int. Conf. on Computer Vision (ICCV)}, 2015.

\bibitem{kollias2021analysing}
Dimitrios Kollias, Irene Kotsia, Elnar Hajiyev, and Stefanos Zafeiriou.
\newblock Analysing affective behavior in the second abaw2 competition.
\newblock {\em arXiv preprint arXiv:2106.15318}, 2021.

\bibitem{kollias2020analysing}
D Kollias, A Schulc, E Hajiyev, and S Zafeiriou.
\newblock Analysing affective behavior in the first {ABAW} 2020 competition.
\newblock In {\em 15th IEEE Int. Conf. on Automatic Face and Gesture
  Recognition (FG)}, 2020.

\bibitem{kollias2019face}
Dimitrios Kollias, Viktoriia Sharmanska, and Stefanos Zafeiriou.
\newblock Face behavior a la carte: Expressions, affect and action units in a
  single network.
\newblock {\em arXiv preprint arXiv:1910.11111}, 2019.

\bibitem{kollias2021distribution}
Dimitrios Kollias, Viktoriia Sharmanska, and Stefanos Zafeiriou.
\newblock Distribution matching for heterogeneous multi-task learning: A
  large-scale face study.
\newblock {\em arXiv preprint arXiv:2105.03790}, 2021.

\bibitem{kollias2019deep}
Dimitrios Kollias, Panagiotis Tzirakis, Mihalis~A Nicolaou, Athanasios
  Papaioannou, Guoying Zhao, Bj{\"o}rn Schuller, Irene Kotsia, and Stefanos
  Zafeiriou.
\newblock Deep affect prediction in-the-wild: {Aff-Wild} database and
  challenge, deep architectures, and beyond.
\newblock {\em Int. Journal of Computer Vision (IJCV)}, pages 1--23, 2019.

\bibitem{kollias2019expression}
Dimitrios Kollias and Stefanos Zafeiriou.
\newblock Expression, affect, action unit recognition: {Aff-Wild2}, multi-task
  learning and {ArcFace}.
\newblock {\em arXiv preprint arXiv:1910.04855}, 2019.

\bibitem{kollias2021affect}
Dimitrios Kollias and Stefanos Zafeiriou.
\newblock Affect analysis in-the-wild: Valence-arousal, expressions, action
  units and a unified framework.
\newblock {\em arXiv preprint arXiv:2103.15792}, 2021.

\bibitem{kossaifi2017afew}
Jean Kossaifi, Georgios Tzimiropoulos, Sinisa Todorovic, and Maja Pantic.
\newblock {AFEW-VA} database for valence and arousal estimation in-the-wild.
\newblock {\em Image and Vision Computing}, 65:23--36, 2017.

\bibitem{kosti2019context}
Ronak Kosti, Jose~M Alvarez, Adria Recasens, and Agata Lapedriza.
\newblock {Context based emotion recognition using {EMOTIC} dataset}.
\newblock {\em IEEE Trans. on Pattern Analysis and Machine Intelligence
  (TPAMI)}, 42:2755--2766, 2019.

\bibitem{kuhnke2020two}
Felix Kuhnke, Lars Rumberg, and J{\"o}rn Ostermann.
\newblock Two-stream aural-visual affect analysis in the wild.
\newblock In {\em 15th IEEE Int. Conf. on Automatic Face and Gesture
  Recognition (FG)}. IEEE, 2020.

\bibitem{liu2020emotion}
Hanyu Liu, Jiabei Zeng, Shiguang Shan, and Xilin Chen.
\newblock Emotion recognition for in-the-wild videos.
\newblock {\em arXiv preprint arXiv:2002.05447}, 2020.

\bibitem{Luo2019ARBEETA}
Y. Luo, J. Ye, R.~B. Adams, J. Li, M.~G. Newman, and J.~Z. Wang.
\newblock {ARBEE}: Towards automated recognition of bodily expression of
  emotion in the wild.
\newblock {\em {Int. Journal of Computer Vision (IJCV)}}, 128:1--25, 2019.

\bibitem{matsumoto1992more}
David Matsumoto.
\newblock More evidence for the universality of a contempt expression.
\newblock {\em Motivation and Emotion}, 16:363--368, 1992.

\bibitem{mittal2020emoticon}
T. Mittal, P. Guhan, U. Bhattacharya, R. Chandra, A. Bera, and D. Manocha.
\newblock {EmotiCon: Context-aware multimodal emotion recognition using Frege's
  principle}.
\newblock In {\em {Proc. IEEE/CVF Conf. on Computer Vision and Pattern
  Recognition (CVPR)}}, 2020.

\bibitem{mollahosseini2017affectnet}
A. Mollahosseini, B. Hasani, and M.~H. Mahoor.
\newblock Affect{N}et: A database for facial expression, valence, and arousal
  computing in the wild.
\newblock {\em {IEEE Trans. Affective Computing}}, 10:18--31, 2017.

\bibitem{mollahosseini2016facial}
Ali Mollahosseini, Behzad Hasani, Michelle~J Salvador, Hojjat Abdollahi, David
  Chan, and Mohammad~H Mahoor.
\newblock Facial expression recognition from world wild web.
\newblock In {\em Proc. IEEE Conf. on Computer Vision and Pattern Recognition
  Workshops (CVPRW)}, 2016.

\bibitem{nicolaou2011continuous}
Mihalis~A Nicolaou, Hatice Gunes, and Maja Pantic.
\newblock Continuous prediction of spontaneous affect from multiple cues and
  modalities in valence-arousal space.
\newblock {\em IEEE Trans. on Affective Computing}, 2:92--105, 2011.

\bibitem{pantic2005web}
Maja Pantic, Michel Valstar, Ron Rademaker, and Ludo Maat.
\newblock Web-based database for facial expression analysis.
\newblock In {\em IEEE Int. Conf. on Multimedia and Expo (ICME)}, 2005.

\bibitem{pennington2014glove}
J. Pennington, R. Socher, and C.~D. Manning.
\newblock Glo{V}e: Global vectors for word representation.
\newblock In {\em {Proc. 2014 Conf. Empirical Methods Natural Language
  Processing (EMNLP)}}, 2014.

\bibitem{pikoulis2021leveraging}
Ioannis Pikoulis, Panagiotis~P Filntisis, and Petros Maragos.
\newblock Leveraging semantic scene characteristics and multi-stream
  convolutional architectures in a contextual approach for video-based visual
  emotion recognition in the wild.
\newblock {\em arXiv preprint arXiv:2105.07484}, 2021.

\bibitem{russell1980circumplex}
James~A Russell.
\newblock A circumplex model of affect.
\newblock {\em Journal of Personality and Social Psychology}, 39(6):1161, 1980.

\bibitem{thinh2021emotion}
Phan Tran~Dac Thinh, Hoang~Manh Hung, Hyung-Jeong Yang, Soo-Hyung Kim, and
  Guee-Sang Lee.
\newblock Emotion recognition with incomplete labels using modified multi-task
  learning technique.
\newblock {\em arXiv preprint arXiv:2107.04192}, 2021.

\bibitem{zafeiriou2017aff}
Stefanos Zafeiriou, Dimitrios Kollias, Mihalis~A. Nicolaou, Athanasios
  Papaioannou, Guoying Zhao, and Irene Kotsia.
\newblock {Aff-Wild}: Valence and arousal 'in-the-wild' challenge.
\newblock In {\em Proc. of the IEEE Conf. on Computer Vision and Pattern
  Recognition Workshops (CVPRW)}, 2017.

\bibitem{zhang2021prior}
Wei Zhang, Zunhu Guo, Keyu Chen, Lincheng Li, Zhimeng Zhang, and Yu Ding.
\newblock Prior aided streaming network for multi-task affective recognitionat
  the 2nd abaw2 competition.
\newblock {\em arXiv preprint arXiv:2107.03708}, 2021.

\bibitem{zhang2020m}
Yuan-Hang Zhang, Rulin Huang, Jiabei Zeng, and Shiguang Shan.
\newblock {$M^{3}T$}: Multi-modal continuous valence-arousal estimation in the
  wild.
\newblock In {\em 15th IEEE Int. Conf. on Automatic Face and Gesture
  Recognition (FG)}. IEEE, 2020.

\bibitem{zhao2016peak}
Xiangyun Zhao, Xiaodan Liang, Luoqi Liu, Teng Li, Yugang Han, Nuno Vasconcelos,
  and Shuicheng Yan.
\newblock Peak-piloted deep network for facial expression recognition.
\newblock In {\em Eur. Conf. on Computer Vision (ECCV)}. Springer, 2016.

\bibitem{zhou2016places}
Bolei Zhou, Agata Lapedriza, Aditya Khosla, Aude Oliva, and Antonio Torralba.
\newblock Places: A 10 million image database for scene recognition.
\newblock {\em IEEE Trans. on Pattern Analysis and Machine Intelligence
  (TPAMI)}, 40:1452--1464, 2018.

\end{thebibliography}
}

\end{document}